\documentclass{ecai} 

\usepackage{latexsym}
\usepackage{amssymb}
\usepackage{amsmath}
\usepackage{amsthm}
\usepackage{booktabs}
\usepackage{enumitem}
\usepackage{graphicx}
\usepackage{color}
\usepackage{hyperref}

\usepackage{paralist}

\usepackage{booktabs}
\usepackage{multicol}
\usepackage{multirow}
\usepackage{bm}
\usepackage{bbm}
\usepackage{diagbox}
\usepackage{makecell}

\newcommand{\BibTeX}{B\kern-.05em{\sc i\kern-.025em b}\kern-.08em\TeX}

\begin{document}


\begin{frontmatter}


\paperid{215} 

\title{TEOcc: Radar-camera Multi-modal Occupancy Prediction via Temporal Enhancement}


\author[A]{\fnms{Zhiwei}~\snm{Lin}\footnote{Equal contribution.}}
\author[A]{\fnms{Hongbo}~\snm{Jin}\footnotemark\footnote{This work was done as an intern at PKU.}}
\author[A]{\fnms{Yongtao}~\snm{Wang}\thanks{Corresponding Author. Email: wyt@pku.edu.cn}} 
\author[B]{\fnms{Yufei}~\snm{Wei}} 
\author[B]{\fnms{Nan}~\snm{Dong}} 

\address[A]{Wangxuan Institute of Computer Technology, Peking University}
\address[B]{Chongqing Changan Automobile Co., Ltd.}


\begin{abstract}
As a novel 3D scene representation, semantic occupancy has gained much attention in autonomous driving.
However, existing occupancy prediction methods mainly focus on designing better occupancy representations, such as tri-perspective view or neural radiance fields, while ignoring the advantages of using long-temporal information.
In this paper, we propose a radar-camera multi-modal temporal enhanced occupancy prediction network, dubbed TEOcc.
Our method is inspired by the success of utilizing temporal information in 3D object detection.
Specifically, we introduce a temporal enhancement branch to learn temporal occupancy prediction.
In this branch, we randomly discard the $t-k$ input frame of the multi-view camera and predict its 3D occupancy by long-term and short-term temporal decoders separately with the information from other adjacent frames and multi-modal inputs.
Besides, to reduce computational costs and incorporate multi-modal inputs, we specially designed 3D convolutional layers for long-term and short-term temporal decoders.
Furthermore, since the lightweight occupancy prediction head is a dense classification head, we propose to use a shared occupancy prediction head for the temporal enhancement and main branches.
It is worth noting that the temporal enhancement branch is only performed during training and is discarded during inference.
Experiment results demonstrate that TEOcc achieves state-of-the-art occupancy prediction on nuScenes benchmarks. 
In addition, the proposed temporal enhancement branch is a plug-and-play module that can be easily integrated into existing occupancy prediction methods to improve the performance of occupancy prediction.
The code and models will be released at \url{https://github.com/VDIGPKU/TEOcc}.

\end{abstract}

\end{frontmatter}


\section{Introduction}

Three-dimensional occupancy prediction is a novel and important task for modern autonomous driving perception systems~\cite{tian2023occ3d}.
Compared to common 3D object detection, occupancy can represent objects in arbitrary shapes with continuous 3D grid cells and semantic labels.
Thus, it can provide fine-grained geometry details, including the specific shape of the foreground object and the concrete geometry of the surrounding background in the whole scene for better perception~\cite{shi2023gridsurvey}.
Besides, it is not insurance enough for autonomous driving scenarios to recognize all predefined objects encountered during training~\cite{boeder2024occflownet}. 
Unseen objects may appear on the road and collide with the self-driving vehicle.
In this situation, 3D occupancy can present novel objects with non-empty grid cells and the 'others' category, avoiding collision.

\begin{figure}[t]
    \centering
\includegraphics[width=0.48\textwidth]{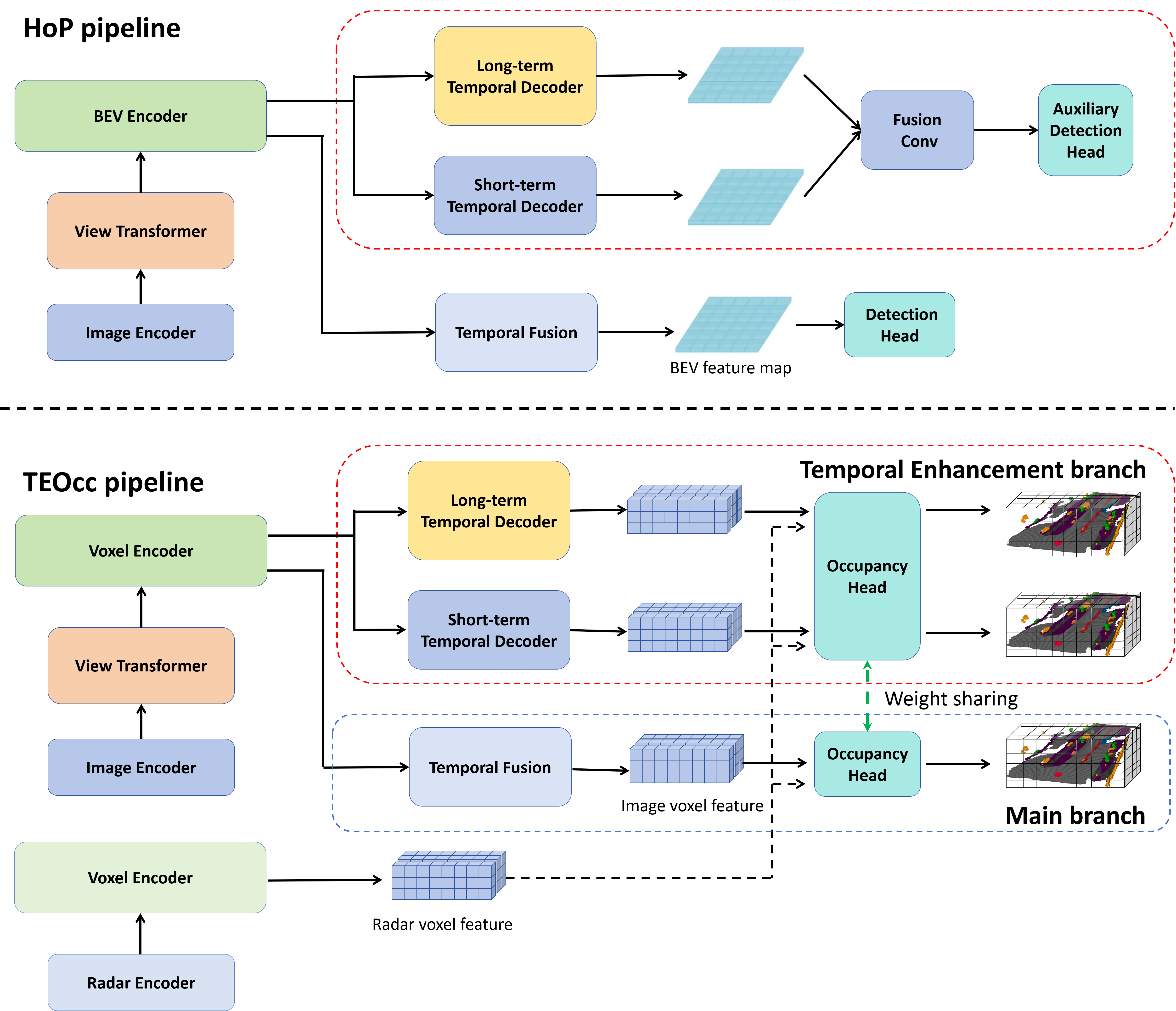}
\vspace{0.5pt}
\caption{\textbf{Differences between HoP and the proposed TEOcc.} TEOcc uses independent long-term and short-term temporal decoders for 3D voxel feature generation and a shared head for occupancy prediction. Besides, TEOcc can incorporate radar-camera multi-modal inputs.}
\vspace{15pt}
\label{fig:fig1}
\end{figure}

Current multi-view camera-based occupancy prediction methods mainly focus on how to represent occupancy, including voxels~\cite{zhang2023occformer}, bird's-eye-view (BEV)~\cite{hou2024fastocc}, tri-perspective view~\cite{huang2023triperspective}, and neural radiance fields (NeRF)~\cite{pan2024renderocc}.
Some of them use hierarchical representations to obtain fine-grained occupancy features from the coarse features~\cite{tian2023occ3d}. 
Although numerous occupancy prediction methods~\cite{cao2022monoscene, tian2023occ3d, huang2023triperspective, li2023fbocc, peng2023learning} have been proposed, they have little exploration in long-term temporal modeling, which achieves great success in 3D object detection~\cite{huang2022bevdet4d, li2022bevstereo, lin2023sparse4d, park2022time, wang2022sts}.
For instance, HoP~\cite{zong2023temporal} is an effective temporal modeling technique in 3D object detection. 
It generates a pseudo Bird’s-Eye View feature of timestamp $t-k$ from its adjacent frames and utilizes this feature to predict the object set at timestamp $t-k$ in the training stage.
However, HoP is only designed for BEV-based camera-only 3D object detection methods.
It cannot be directly applied to the multi-modal occupancy prediction task because of different feature representations and complex multi-modal inputs.

\begin{figure*}[!th]
    \centering
\includegraphics[width=17cm]{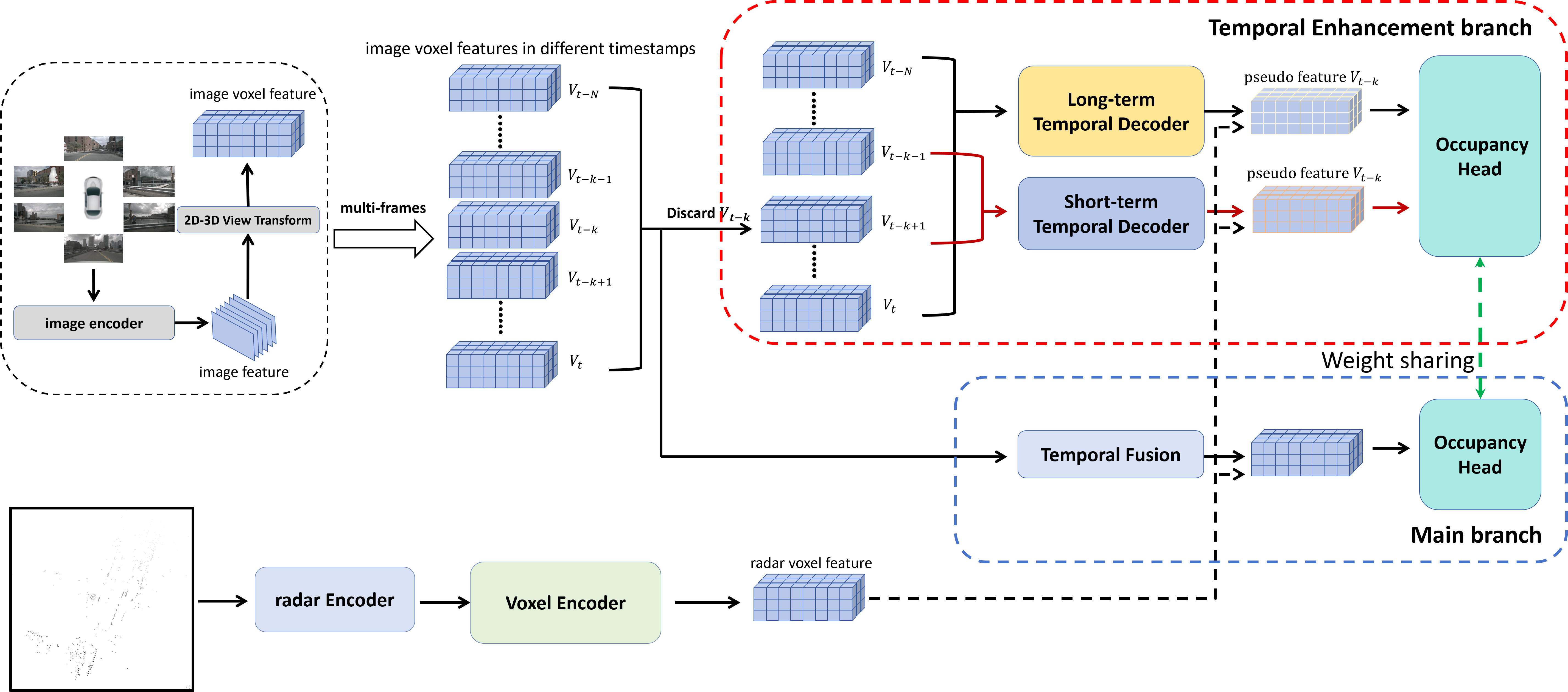}
\caption{\textbf{Overall pipeline of TEOcc}. First, multi-frame multi-view camera features are extracted with an image encoder.
The extracted 2D image features are transformed into 3D image voxel features with a 2D-3D view transformation module. 
Parallelly, we use a radar encoder and voxel encoder to extract radar voxel features.
After that, in the main branch, all temporal image voxel features and radar voxel features are kept to predict final occupancy results.
In the temporal enhancement branch, we discard one image voxel feature and use long-term and short-term decoders to generate corresponding pseudo features.
Finally, a shared occupancy head is used to predict occupancy from generated pseudo voxel features.
}
\label{fig:pipeline}
\vspace{15pt}
\end{figure*}

To this end, we propose a radar-camera multi-modal occupancy prediction network with a temporal enhancement branch, named TEOcc, 
Specifically, in the temporal enhancement branch, we randomly mask one frame of the temporal multi-view camera inputs and generate its pseudo features with a long-term and short-term temporal decoder.
Note that we only perform the temporal enhancement branch during training. 
Thus, no extra overheads are introduced during inference.
Different from HoP, to reduce training costs and combine radar features, we design long-term and short-term temporal decoders with 3D convolutional layers.
Besides, HoP concatenates features of long-term and short-term temporal decoders to produce one pseudo feature, while TEOcc uses independent temporal decoders to generate two pseudo features.
Furthermore, we propose to use a shared 3D occupancy prediction head for two pseudo features and the main branch to predict corresponding occupancy results.
The reason is that the occupancy prediction head is more like a dense classification head, which maps voxel features to semantic results. Thus, using a shared occupancy prediction head for all features can learn a better mapping.

The main contributions of this work are summarized as follows:
\vspace{5pt}
\begin{compactitem}
    \item 
    We propose TEOcc, a radar-camera multi-modal temporal enhanced occupancy prediction network, which extends HoP to the multi-modal occupancy prediction task.
    \vspace{3pt}
    \item 
    Different from HoP, we use hand-craft long-term and short-term temporal decoders to independently predict occupancy with a shared occupancy prediction head.
    \vspace{3pt}
    \item
    Building upon the current competitive 3D occupancy prediction method, TEOcc achieves state-of-the-art 3D occupancy estimation performance on the nuScenes dataset.
\end{compactitem}

\section{Related Works}
\subsection{3D Occupancy Prediction}

The ability to forecast 3D occupancy, creating a comprehensive 3D voxel-based semantic depiction of a scene, is demanding and suited for autonomous driving systems.
The emergence of Occ3D-nuScenes dataset \cite{tian2023occ3d} provided a fine-grained 3D occupancy ground truth and sparked more exploration in this field.
Recently, several occupancy prediction methods have been proposed.

For point-based methods,
S3CNet \cite{cheng2020s3cnet} formulates a sparse convolution-based neural network that deals with the sparsity of large-scale environments and predicts the semantically completed scene from the LiDAR point cloud.
AIC-Net \cite{li2020anisotropic} presents an anisotropic convolutional network with adaptability to dimensional anisotropy and implicitly enables 3D kernels with varying sizes.
Local-DIFs \cite{rist2021semantic} produces a representation for 3D scenes by deep implicit functions with spatial support and generates point-like training targets from LiDAR data.
More recently, PointOcc \cite{zuo2023pointocc} introduces a cylindrical tri-perspective view to represent point clouds for occupancy prediction.
OccWorld \cite{zheng2023occworld} proposes to learn the movement of the ego car and the evolution of the surrounding scenes simultaneously with a world model.

\begin{figure*}[t]
    \centering
\includegraphics[width=12cm]{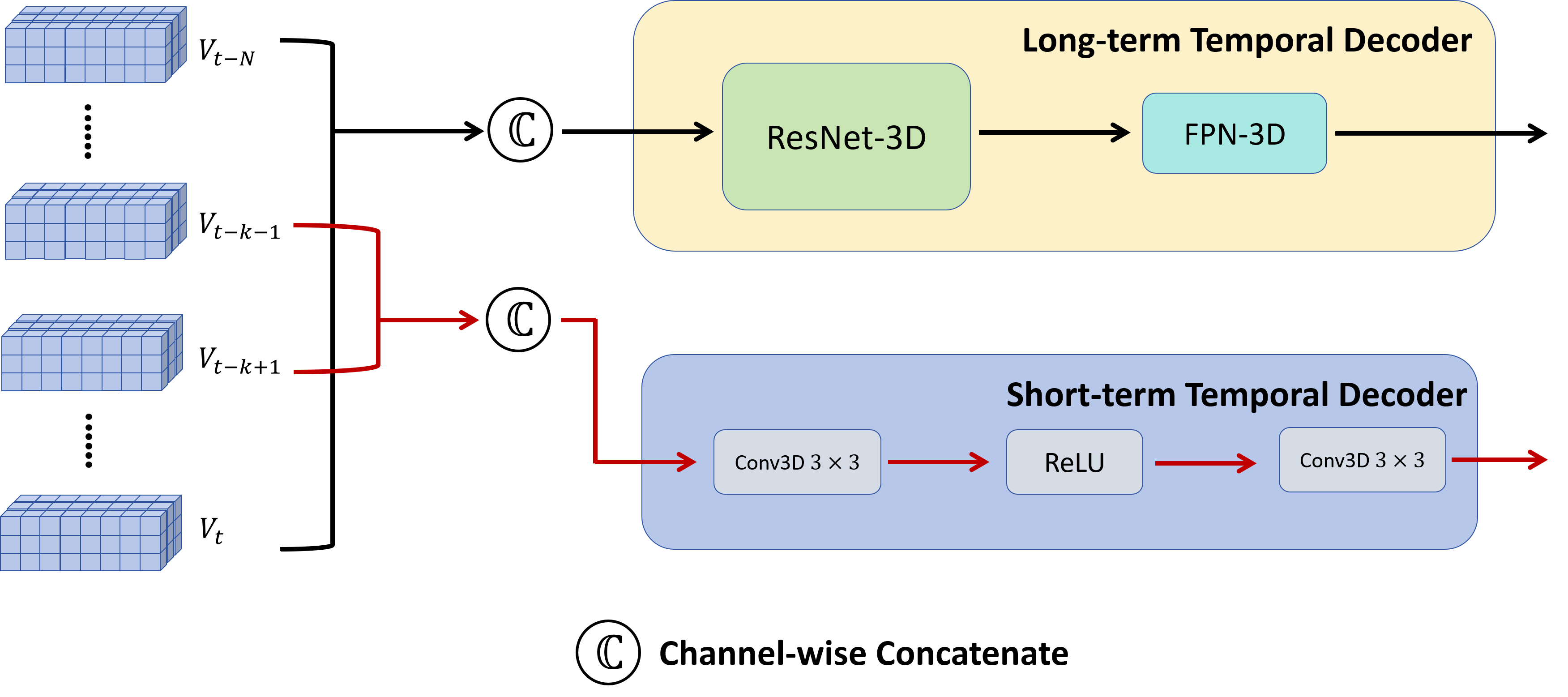}
\caption{\textbf{Architecture of the temporal enhancement module}.
The long-term temporal decoder consists of a ResNet-3D backbone and a FPN-3D neck to process multi-scale 3D voxel features.
The short-term decoder is composed of two 3D convolution layers. 
}
\label{fig:temporal}
\vspace{15pt}
\end{figure*}

For camera-based methods, MonoScene \cite{cao2022monoscene} utilizes 3D Context Relation Prior and 2D-3D U-nets to enhance spatial-semantic awareness.
SelfOcc \cite{huang2023selfocc} explores a self-supervised way to learn 3D occupancy using only video sequences.
OccFormer \cite{zhang2023occformer} designs a dual-path transformer network to effectively process the 3D volume for semantic occupancy prediction.
FB-OCC \cite{li2023fbocc} proposes a bidirectional projection framework to utilize both forward and backward projection and avoid their limitations, obtaining better dense representation.
TPVFormer \cite{huang2023triperspective}
introduces a tri-perspective view as a new representation to depict the 3D scenes for predicting semantic occupancy.
SurroundOcc \cite{wei2023surroundocc} designs a pipeline to generate dense occupancy ground truth by fusing multi-frame LiDAR scans of dynamic objects and static scenes separately.
RenderOcc \cite{pan2024renderocc} attempts to train multi-view 3D occupancy networks solely using 2D labels via volume rendering.
Univision \cite{hong2024univision} simultaneously handles occupancy prediction and object detection utilizing a unified representation. 
FastOcc \cite{hou2024fastocc} accelerates the model while keeping its accuracy by replacing the time-consuming 3D convolution network with a novel residual architecture, where features are mainly extracted by a lightweight 2D BEV convolution network.

In contrast, our proposed TEOcc introduces temporal information in occupancy prediction via temporal enhancement and presents a radar-camera multi-modal occupancy prediction network.

\subsection{Temporal Modeling in 3D Object Detection}
Motion information has been extensively explored to improve performance with temporal cues in the 3D object detection task.
BEVFormer \cite{li2022bevformer} designs the temporal self-attention mechanism to dynamically fuse the previous BEV features by deformable attention \cite{zhu2021deformable} in an RNN manner. 
BEVDet4D \cite{huang2022bevdet4d} introduces the temporal modeling to lift BEVDet \cite{huang2022bevdet} to spatial-temporal 4D space.
BEVStereo \cite{li2022bevstereo} proposes a dynamic temporal stereo technique for tackling the ill-posed issue of depth perception.
STS \cite{wang2022sts} brings the temporal stereo technique in multi-view 3D object detection to facilitate accurate depth learning and 3D detection. 
SOLOFusion \cite{park2022time} utilizes a multi-view stereo (MVS) method to process high-resolution short-term images and then warp low-resolution long-term BEV features to produce a fused temporal BEV feature.
PETRv2 \cite{liu2022petrv2} extends the position embedding transformation to temporal representation learning.
HoP \cite{zong2023temporal} encourages more accurate BEV feature learning via performing object detection in the historical frame.
More recently, StreamPETR \cite{wang2023exploringStreamPETR} proposes to utilize sparse object queries as intermediate representations to capture
temporal information.
SparseBEV \cite{Liu_2023_ICCVSparseBEV}
incorporates an adaptive spatio-temporal sampling module to perceive the BEV and temporal information dynamically.

These methods show great effectiveness for 3D object detection.
However, few attempts have been made to explore temporal enhancement techniques for better perception performance in the 3D occupancy prediction task.
Our proposed TEOcc tries to address this deficiency.


\section{Method}

\subsection{Overall Architecture}
Our overall architecture is shown in Figure \ref{fig:pipeline}.
Specifically, we first extract multi-frame 3D volume features with the image encoder and view-transformation module.
Meanwhile, we extract radar voxel features with a radar encoder.
Then, in the temporal enhancement branch, we discard the image feature of the selected frame and reconstruct it using features from other frames and radar.
The reconstructed pseudo feature and current 3D feature are sent to a shared occupancy head for final occupancy and semantic prediction.

\subsection{Temporal Enhancement Branch}
\label{sec:3.2}
As shown in Figure \ref{fig:temporal}, our temporal enhancement branch comprises independent long-term and short-term temporal decoders.
Specifically, given the 3D image voxel feature sequence $\{{V}_{t-N}, {V}_{t-N+1},..., {V}_{t}\}$ that consists of $N$ historical image voxel features and the current image voxel feature, we randomly mask out the 3D image voxel feature $V_{t-k}$.
Then, we send the remaining 3D image voxel feature sequence $\{V_{t-N},..., {V}_{t}\} - \{V_{t-k}\}$ and radar voxel feature to long-term temporal decoder.
For short-term temporal decoder, we use adjacent features $\{{V}_{t-k-1}, {V}_{t-k+1}\}$ and radar voxel feature as the input.
Two temporal decoders predict two pseudo-3D voxel features ${V}_{t-k}$.
Finally, we use the occupancy head to predict the occupancy results for the $t-k$ frame.

\begin{figure}[!t]
    \centering
\includegraphics[width=0.9\linewidth]{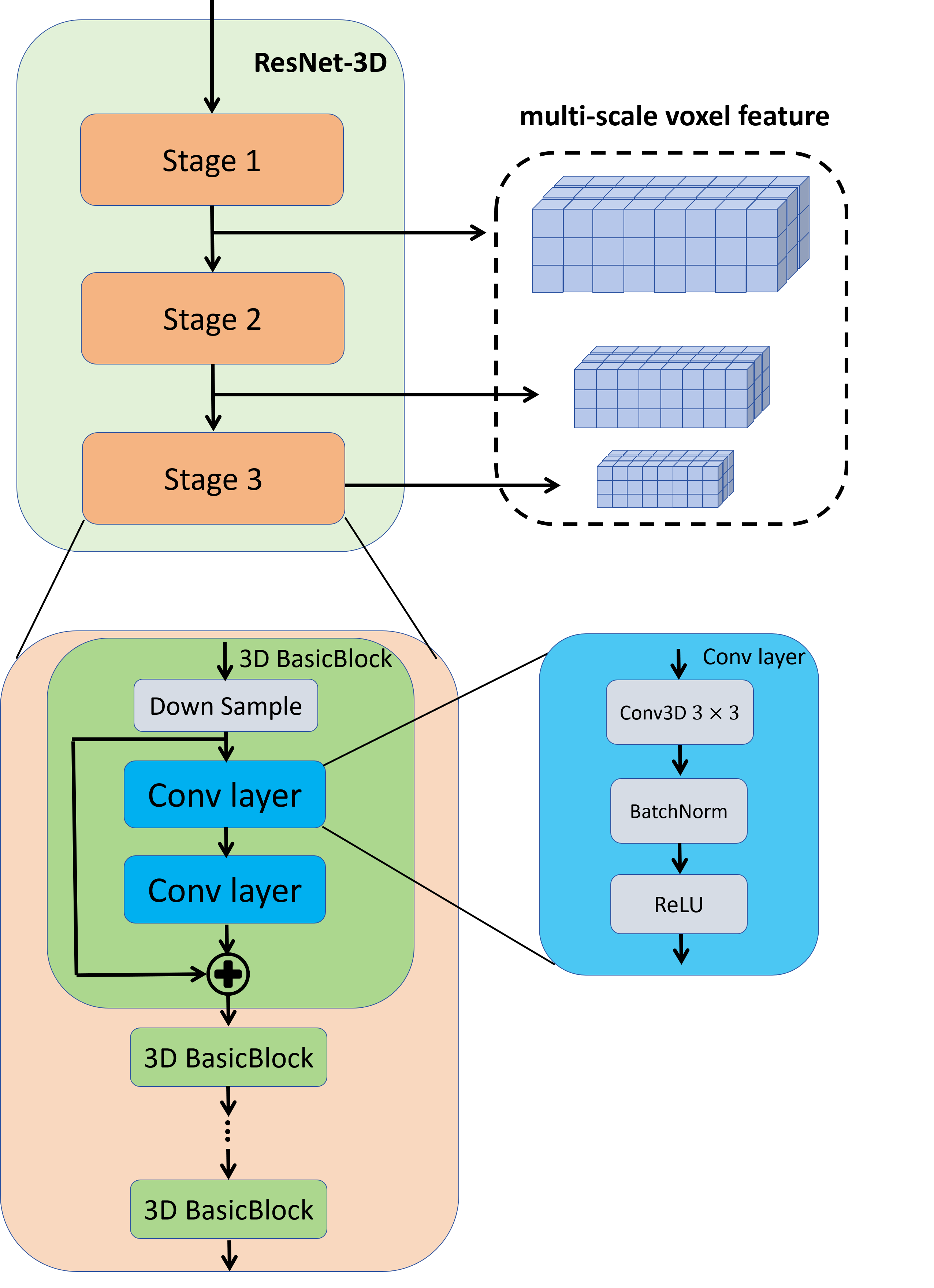}
\caption{\textbf{Architecture of ResNet-3D}. ResNet-3D has three stages. Each stage consists of several 3D BasicBlocks.
}
\label{fig:resnet}
\vspace{20pt}
\end{figure}

\vspace{3pt}
\noindent\textbf{Temporal Decoder.}
We designed two different temporal decoders based on the different temporal inputs following HoP.
The short-term temporal decoder mainly focuses on the information from adjacent voxel features set $\{V_{t-k-1}, V_{t-k+1}\}$, while the long-term temporal decoder processes the whole temporal voxel feature set $\{V_{t-N},..., V_{t-k-1}, V_{t-k+1},..., V_{t-1}\}$.
Due to the high temporal correlation between adjacent frames, the short-term temporal decoder can create a detailed spatial representation for $V_{t-k}$.
In contrast, the long-term temporal decoder perceives the motion clues over long-term history, which improves the localization accuracy~\cite{park2022time}. 
Therefore, these two branches are complementary to each other.

As depicted in HoP~\cite{zong2023temporal}, Deformable Attention shows powerful historical object prediction ability because the coordinate offsets in Deformable Attention can match the movement of foreground objects and model temporal motion cues for the 3D object detection task.
However, in the occupancy prediction task, every voxel needs to be classified rather than only predicting foreground objects.
Therefore, Deformable Attention may not be appropriate for the 3D occupancy prediction task.
Besides, full 3D Attention can capture all voxels in 3D space, but resulting in huge computational costs and time overhead.

Fortunately, in our experiments, we find that simple 3D convolutions not only capture the movement of objects with temporal information, but also obtain a precise dense voxel representation for the 3D occupancy prediction task.
Therefore, we replace Deformable Attention in HoP for temporal decoders with 3D convolution and specially design convolutional layers.

\begin{figure}[!t]
    \centering
\includegraphics[width=0.9\linewidth]{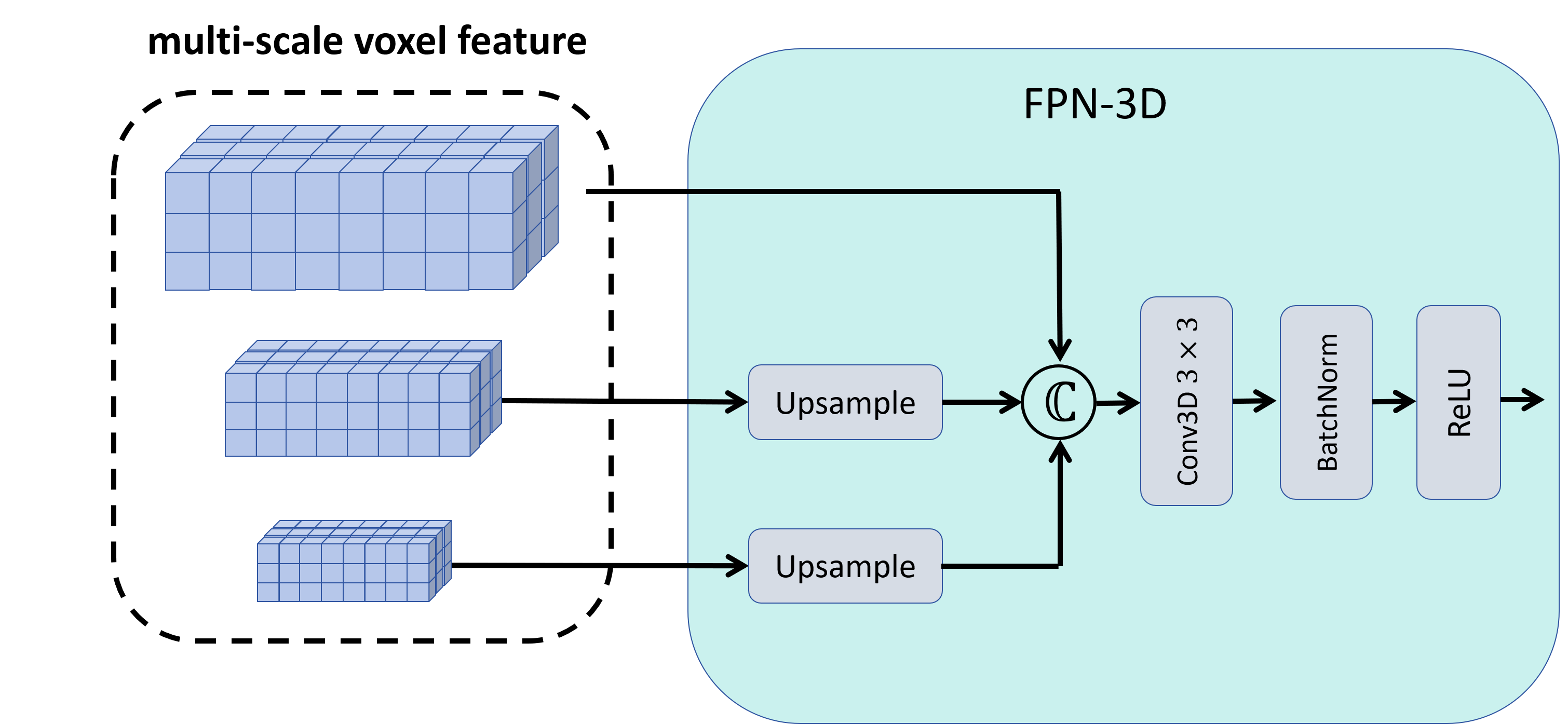}
\caption{\textbf{Architecture of FPN-3D}. We upsample multi-scale voxel features into one scale and fuse them with a 3D convolution layer.
}
\vspace{10pt}
\label{fig:fpn}
\end{figure}

Specifically, as shown in Figure~\ref{fig:temporal}, we design ResNet-3D and FPN-3D as long-term temporal decoders.
More concretely, ResNet-3D consists of three stages with the downsampling operation. 
Each stage is composed of several 3D BasicBlocks.
Every 3D BasicBlocks has two 3D convolutional layers followed by a ReLU activation layer, as illustrated in Figure~\ref{fig:resnet}.
In the three stages of ResNet-3D, we obtain three 3D voxel features in different scales.
To further fuse these 3D voxel features with different resolutions, we send them to the FPN-3D.
As shown in Figure~\ref{fig:fpn}, we first use trilinear interpolation to upsample these 3D voxel features to one resolution. Finally, we concatenate the upsampled 3D voxel features and send them to a convolutional layer followed by a norm and a ReLU activation layer.
Using this multi-scale feature pyramid network to process image features, we can enhance spatial recognition ability and strong adaptability for the dense occupancy prediction task.

For short-term temporal decoders, since there is a high temporal correlation between voxel features from two adjacent frames, we find that two 3D convolutional layers with a ReLU activate layer can fuse the adjacent voxel features well.

\vspace{3pt}
\noindent\textbf{Occupancy head.}
Different from HoP, we do not use an extra auxiliary occupancy head.
In contrast, we use a shared occupancy head with multi-layer perceptions.
The reason is that the occupancy head serves as the mapping function from voxel features to the occupancy category.
Thus, it is not relevant to a specific frame, and the mapping can be learned by the same lightweight occupancy head used in the main branch.

\begin{table*}[!t]
    \centering
    \resizebox{0.98\textwidth}{!}{
    \begin{tabular}{c|c|*{17}{c}}
         \hline
         \multirow{1}{*}[4ex]{\shortstack{Method}}  & \multirow{1}{*}[4ex]{\shortstack{mIoU}} &\rotatebox{90}{Others}&\rotatebox{90}{Barrier}&\rotatebox{90}{Bicycle}&\rotatebox{90}{Bus}&\rotatebox{90}{Car}&\rotatebox{90}{Cons.veh}&\rotatebox{90}{Motorcycle}&\rotatebox{90}{Pedestrian}&\rotatebox{90}{Traffic cone}&\rotatebox{90}{Trailer}&\rotatebox{90}{Truck}&\rotatebox{90}{Dri.sur}&\rotatebox{90}{Other flat}&\rotatebox{90}{Sidewalk}&\rotatebox{90}{Terrain}&\rotatebox{90}{Manmade}&\rotatebox{90}{Vegetation}\\
        \hline
         MonoScene&6.06&1.75&7.23&4.26&4.93&9.38&5.67&3.989&3.01&5.90&4.45&7.17&14.91&6.32&7.92&7.43&1.01&7.65\\
         BEVFormer&26.88 &5.85& 37.83& 17.87& 40.44& 42.43& 7.36& 23.88& 21.81 &20.98& 22.38& 30.70& 55.35& 28.36& 36.00& 28.06& 20.04& 17.69 \\
         BEVStereo&24.50&15.73&38.41&7.88&38.70&41.20&17.56&17.33&14.69&10.31&16.84&29.62&54.08&28.92&32.68&26.54&18.74& 17.49\\
         OccFormer&21.93&5.94&30.29&12.32&34.40&39.17&14.44&16.45&17.22&9.27&13.90&26.36&50.99&30.96&34.66&22.73&6.76&6.97\\
         RenderOcc&26.11&4.84&31.72&10.72&27.67&26.45&13.87&18.2&17.67&17.84&21.19&23.25&63.2&36.42&46.21&44.26&19.58&20.72\\

        
        TPVFormer&27.83&7.2&38.9&13.7&40.8&45.9&17.2&20.0&18.8&14.3&26.7&34.2&55.6&35.5&37.6&30.7&19.4&16.78\\

         SurroundOcc&37.18& 8.97&46.33 &17.08& 46.54& 52.01 &20.05& 21.47& 23.52& 18.67& 31.51& 37.56& 81.91& 41.64& 50.76& 53.93& 42.91&37.16\\
         FB-Occ&39.11&13.57& 44.74& 27.01& 45.41& 49.10& 25.15& 26.33& 27.86& 27.79& 32.28 &36.75& 80.07& 42.76& 51.18& 55.13& 42.19& 37.53\\
         FastOcc &39.21&12.06 &43.53 &28.04& 44.80& 52.16& 22.96& 29.14& 29.68& 26.98& 30.81& 38.44& 82.04& 41.93& 51.92& 53.71 &41.04&35.49\\
         \hline
         TEOcc (Ours)&39.36&9.59&47.60&13.82&43.91&52.87&27.92&17.58&23.89&21.69&33.91&39.60&83.38&41.84&54.94&57.92&50.83&47.23\\
         TEOcc-RC (Ours)&42.90&10.82&50.33&24.28&48.99&57.32&29.38&24.41&30.14&28.46&36.46&43.01&83.96&43.09&56.00&59.34&54.18&49.16\\
         \hline
    \end{tabular}
    }
    \vspace{5pt}
    \caption{\textbf{Comparison of 3D occupancy prediction results on OCC3D-nuScenes val set.} `Cons.veh' and `Dri.sur' are the shorts for construction vehicles and driveable surfaces. `TEOcc-RC' means the radar-camera multi-modal version of TEOcc.}
    \vspace{10pt}
    \label{tab:table1}
\end{table*}

\subsection{Radar-camera Fusion with Temporal Enhancement}
In Sections~\ref{sec:3.2}, we discuss the temporal enhancement for multi-view camera-based 3D occupancy prediction methods. This section extends the temporal enhancement to radar-camera multi-modal fusion 3D occupancy prediction.

Specifically, we follow the pipeline of BEVFusion and replace the unified BEV space with the unified 3D voxel space.
Specifically, as shown in Figure~\ref{fig:pipeline}, we use a radar encoder and a voxel encoder to extract radar voxel features.
Then, radar voxel features are sent to the main and temporal enhancement branches for radar-camera fusion.
Besides, following BEVFusion, we concatenate the image voxel features and radar voxel features and use a 3D convolutional layer followed by a norm and a ReLU activation layer as the fusion layers.
We utilize three independent fusion layers for the main and temporal enhancement branches.
Finally, the fused multi-modal features are sent to the shared occupancy head to predict occupancy results.

\subsection{Training and Inference}
During the training stage, we keep the original occupancy loss of the main branch and add two additional occupancy losses from the temporal enhancement branch.
The overall optimization objective is formulated as follows:
\begin{equation}
    \mathcal{L} = \mathcal{L}_{Occ} + \mathcal{L}_{Occ\_long} + \mathcal{L}_{Occ\_short},
\end{equation}
where $\mathcal{L}_{Occ\_long}$ and $\mathcal{L}_{Occ\_short}$ denote the occupancy loss from long-term and short-term temporal decoders, respectively.

For inference, the temporal enhancement branch is removed. We only use the occupancy prediction from the main branch.
Therefore, no extra inference cost is introduced.

\section{Experiment}

\subsection{Dataset}
We evaluate our method on Occ3D-nuScenes \cite{tian2023occ3d} benchmark. 
The Occ3D-nuScenes dataset is built upon the widely used large-scale autonomous driving dataset, nuScenes \cite{caesar2020nuscenes}, which includes 1000 outdoor driving scenes with six surrounding-view cameras, LiDAR, and radar sensors. 
To provide high-quality occupancy labels, Occ3D-nuScenes first separates dynamic and static objects with LiDAR segmentation labels provided by nuScenes.
Then, it aggregates multi-frame LiDAR points and utilizes the K-Nearest-Neighbor algorithm and mesh reconstruction to obtain a dense voxel with classification labels for occupancy.
Occ3D-nuScenes provides 16 classes and a free class of 3D semantic labels for each scene. 
Each sample covers a range of [-40m, 40m], [-40m, 40m], and [-1m, 5.4m] with a voxel size of 0.4m for the x, y, and z-axis, respectively.
The metric used in this benchmark is the mean Intersection over Union (mIoU) score.
To obtain mIoU, we calculate the IoU value for each class and average the IoU value over 17 classes.

\subsection{Implementation Details}
We use a multi-view camera occupancy prediction method, BEVStereo \cite{li2022bevstereo}, as the baseline for TEOcc.
BEVStereo is composed of an image encoder, a 2D-3D view transformer, a BEV encoder, and an occupancy prediction head.
We maintain the majority of the original structure of BEVStereo and add the proposed temporal enhancement module to construct the multi-view camera-based TEOcc.
Then, we train radar-camera multi-modal TEOcc-RC based on the single-modal TEOcc.

For image feature extraction, we use ResNet50 \cite{he2015deep} and FPN \cite{lin2017feature} as the image encoder.
We employ 9 temporal frames.
The image size is set to 256$\times$704.
For the radar encoder, we employ PointPillar with a voxel size of [0.4, 0.4, 0.4] for the x, y, and z axes.
PointPillar first divides radar points into several pillars according to the voxel size. Then, it uses a simplified version of PointNet to extract features of radar points in each pillar. Finally, the pillar features are scattered to create a 2D Bird’s-Eye View feature.
We use Adam~\cite{DBLP:journals/corr/KingmaB14adam} as the optimizer with a batch size of 4.
We train the network 24 epochs with a learning rate of 1e-4.
For data augmentation, we use the same image augmentation with BEVPoolv2~\cite{huang2022bevpoolv2}, \textit{i.e.}, image rotation and flip.
Besides, we use horizontal flips as the augmentation in voxel space.

\subsection{Main Results}

We compare the proposed TEOcc with previous state-of-the-art 3D occupancy prediction methods on the Occ3D-NuScenes validation set in Table \ref{tab:table1}.
TEOcc shows competitive 3D occupancy prediction performance.
Specifically, TEOcc achieves a mIoU of 39.36, outperforming all previous multi-view camera-based occupancy prediction methods, including TPVFormer, SurroundOcc, and FastOcc.
Besides, the challenge of dynamic object recognition described in RenderOCC is alleviated by our method.
In particular, for dynamic objects like cars, buses, trailers, and trucks, TEOcc significantly improves the occupancy performance compared with RenderOcc. 
In addition, for static objects, TEOcc is still ahead of RenderOcc, demonstrating the effectiveness of temporal enhancement for the comprehensive understanding of 3D spatial relationships.
Furthermore, when combined with radar inputs, TEOcc-RC improves TEOcc by 3.54 mIoU, surpassing the previous state-of-the-art 3D occupancy prediction method FastOcc by 3.69 mIoU.

In summary, the results indicate that TEOcc with temporal enhancement can improve the construction and perception of occupancy representations for existing frameworks and enhance the overall understanding of 3D scenes.

\begin{table}[t]
    \centering
    \begin{tabular}{ccc|c}
         \hline
         long-term&short-term&random&mIoU\\
         \hline
         &&&21.02\\
         \checkmark & & &  24.33 \\
         \checkmark & \checkmark  &&  26.12\\
         \checkmark &\checkmark &  \checkmark & 26.95 \\
         \hline
    \end{tabular}
    \vspace{5pt}
    \caption{\textbf{Ablation of main components.}
    Each component improves the 3D occupancy performance consistently.}
    \vspace{10pt}
    \label{tab:main components}
\end{table}

\begin{table}[t]
    \centering
    \begin{tabular}{ccc|c}
         \hline
         long-term&short term&fusion&mIoU\\
         \hline
         \checkmark & & &  24.33\\
          \checkmark& \checkmark & &26.12\\
         \checkmark& \checkmark & \checkmark &24.65\\
        
         \hline
    \end{tabular}
    \vspace{5pt}
    \caption{\textbf{Ablation of independent temporal decoders.} Using independent temporal decoders achieves better results than fusing 3D voxel features from two decoders.}
    \vspace{10pt}
    \label{tab:independent}
\end{table}

\begin{table}[t]
    \centering
    \begin{tabular}{c|c}
         \hline
         flip augmentation&mIoU\\
         \hline
          &  31.16\\
          \checkmark & 32.05  \\
        
         \hline
    \end{tabular}
    \vspace{5pt}
    \caption{\textbf{Ablation of voxel data augmentation.} Filp data augmentation in voxel space improves occupancy performance.}
    \vspace{10pt}
    \label{tab:augmentation}
\end{table}

\begin{table}[!h]
    \centering
    \begin{tabular}{c|c}
         \hline
         shared occupancy head&mIoU\\
         \hline
          & 24.63 \\
          \checkmark & 26.12 \\
        
         \hline
    \end{tabular}
    \vspace{5pt}
    \caption{\textbf{Ablation of occupancy head.} Sharing an occupancy head obtains better occupancy results.}
    \vspace{7pt}
    \label{tab:head}
\end{table}

\begin{table}[t]
    \centering
    \begin{tabular}{c|cc}
         \hline
         Method & GPU Memory  & Training Time\\
         \hline
          BEVStereo&1$\times$&1$\times$  \\ 
          TEOcc&1.8$\times$&1.1$\times$  \\
         \hline
    \end{tabular}
    \vspace{5pt}
    \caption{\textbf{Ablation of training cost.} TEOcc brings marginal training time and GPU memory increasing.}
    \vspace{10pt}
    \label{tab:time}
\end{table}

\begin{figure*}[!t]
    \centering
\includegraphics[width=17cm]{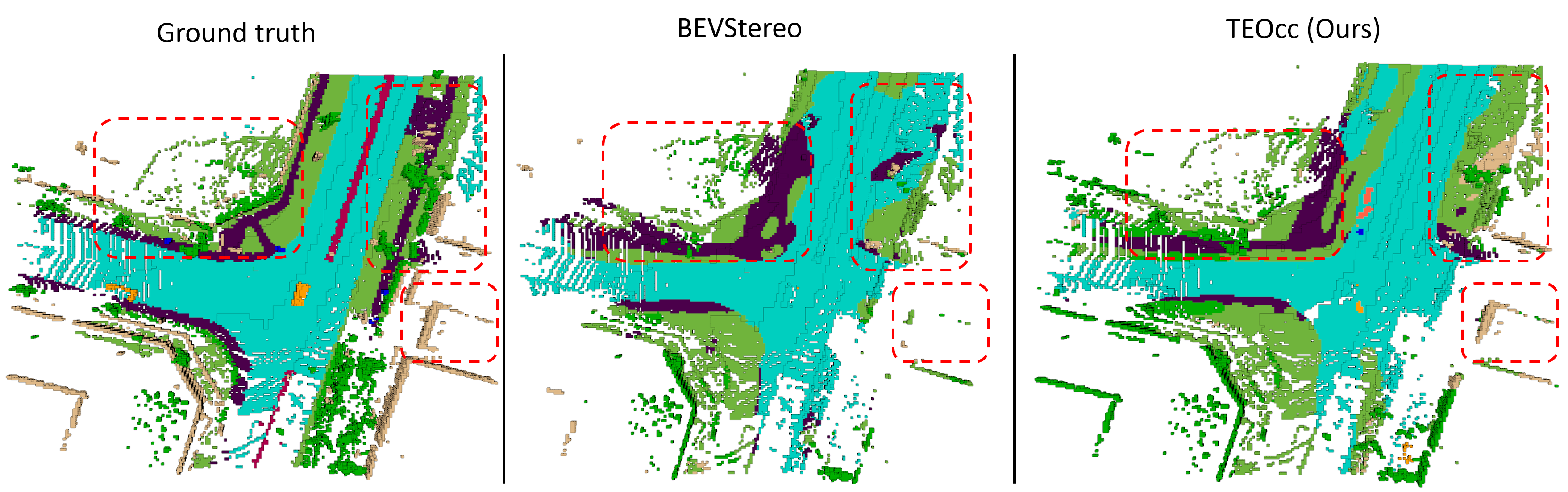}
\vspace{3pt}
\caption{\textbf{Occupancy visualization in global view}. From left to right, we show the ground truth, BEVStereo (baseline), and TEOcc prediction results.
We can see that TEOcc is more accurate in perceiving global details, especially distant information and occluded parts.}
\vspace{15pt}
\label{fig:fig4}
\end{figure*}

\begin{figure*}[!t]
    \centering
\includegraphics[width=17cm]{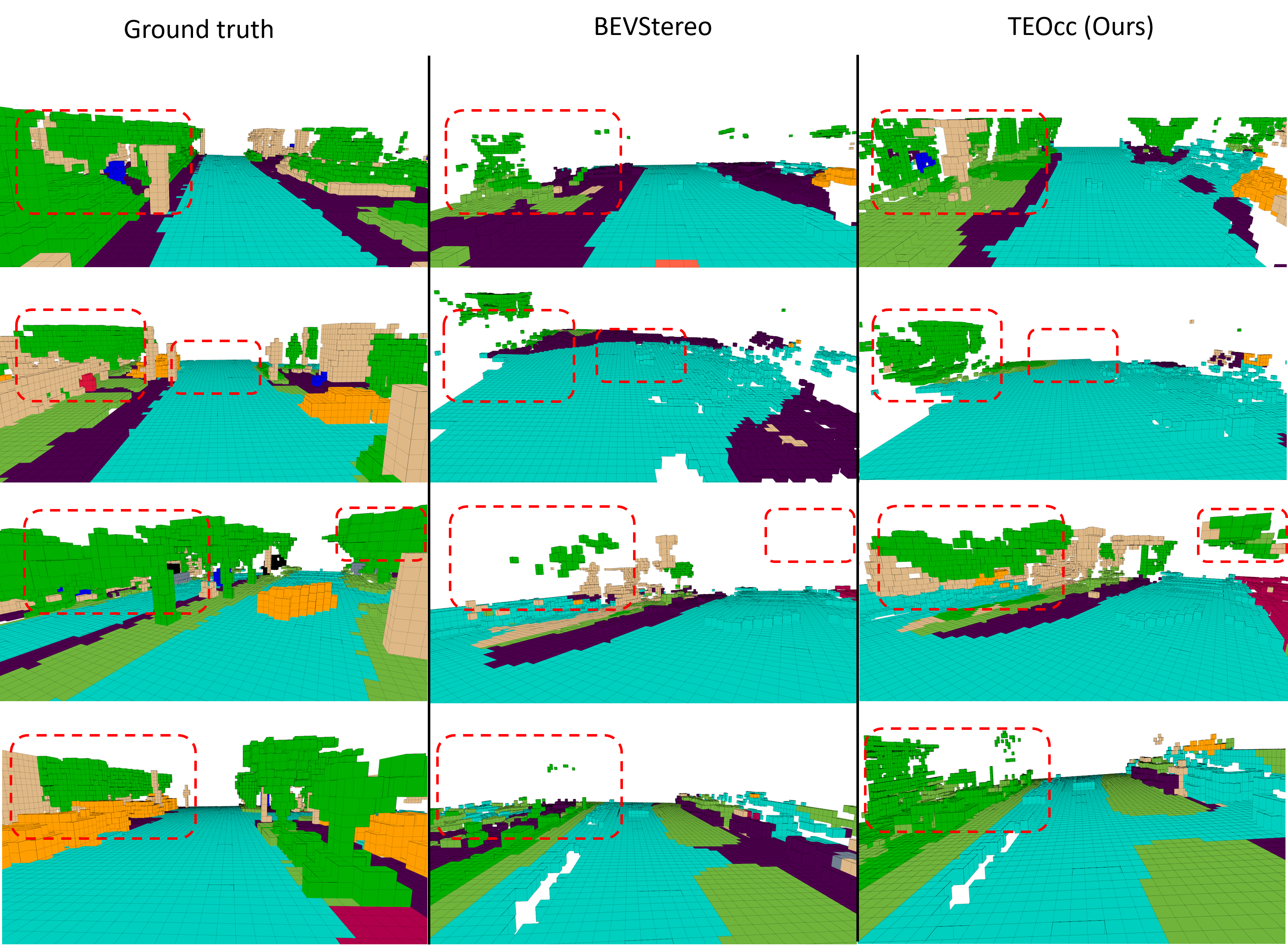}
\vspace{3pt}
\caption{
\textbf{Occupancy visualization in local view.}
From left to right, we show ground truth, BEVStereo (baseline), and TEOcc prediction results.
It can be observed that TEOcc is better at predicting object shape.
}
\vspace{15pt}
\label{fig:fig5}
\end{figure*}

\subsection{Ablation Study}

To validate the effectiveness of the proposed module, we conduct extensive ablation studies on the Occ3D-NuScenes dataset. In this section, to reduce the training costs, we train TEOcc with 0.5$\times$ schedule, \textit{i.e.}, 12 epochs.

\vspace{3pt}
\noindent\textbf{Main Components.} 
We conduct comprehensive experiments to verify the effectiveness of each module.
The main results of the experiment are shown in Table~\ref{tab:main components}.
Long-term enhancement networks provide the most significant performance improvement, with 3.31 mIoU.
The short-term temporal decoder helps to improve the final prediction results with 1.79 mIoU.
It is worth noting that we use a random selection strategy to select which frame is masked in the temporal enhancement branch, while HoP only chooses the $t-1$ frame for masking.
The results show that using the random selection strategy results in a noticeable performance boost.
We speculate that the {random selection strategy} makes the model pay more attention to the temporal changes of surrounding scenes rather than mechanically memorizing the positional relationship between the past and the current frame.

\vspace{3pt}
\noindent\textbf{Independent Temporal Decoders.}
Different from HoP \cite{zong2023temporal}, we obtain two independent 3D voxel features with fine-grained granularity from the two temporal decoders.
As shown in Table~\ref{tab:independent}, using independent temporal decoders shows better occupancy prediction performance compared with fusing the 3D voxel features from two temporal decoders.
We speculate that the long-term and short-term temporal encoders learn different 3D voxel features due to the different temporal lengths. Fusing them into one 3D voxel feature may lead to feature conflicts.

\vspace{3pt}
\noindent\textbf{Data Augmentation.}
In addition to image augmentation, we follow the BEV data augmentation in BEVPoolv2~\cite{huang2022bevpoolv2} to add voxel data augmentation.
As shown in Table~\ref{tab:augmentation}, we find horizontal flip data augmentation in voxel space can improve the occupancy prediction performance from 31.16 mIoU to 32.05 mIoU.

\vspace{3pt}
\noindent\textbf{Shared Occupancy Head.}
As shown in Table~\ref{tab:head}, we compare the results of using the additional auxiliary occupancy head and shared occupancy head for the final prediction.
%
The results show that employing a sharing occupancy head obtains 1.51 mIoU improvement compared with the additional auxiliary head.
The reason is that the occupancy head is a dense classification head, which maps 3D voxel features to the occupancy category.
Thus, the shared occupancy head forces the generated pseudo 3D voxel features to share the same feature space with the main branch, allowing the temporal enhancement branch to learn a more unified temporal representation with smaller training costs. 

\vspace{3pt}
\noindent\textbf{Training Costs.}
Because we add the temporal enhancement branch during the training stage, the training cost increases. 
To evaluate the efficiency of TEOcc in training, we compare its training time and GPU memory consumption with the BEVStereo baseline.
This training time is evaluated with 8 NVIDIA A800 GPUs.
As shown in Table \ref{tab:time}, the additional training time brought by TEOcc is negligible.
However, due to the utilization of 3D convolution in temporal decoders, the memory consumption increases to 1.8$\times$.

\subsection{Visualization}
We provide the occupancy visualization of the global view and local view in Figure \ref{fig:fig4} and \ref{fig:fig5}, respectively.
The figures show that, with the proposed temporal enhancement module, TEOcc can capture more accurate long-distance view information and predict detailed scenes locally.

\section{Conclusion}
In this paper, we propose a radar-camera multi-modal temporal enhanced occupancy prediction network, named TEOcc. 
Specifically, we generate a pseudo voxel feature of timestamp $t-k$ from its adjacent frames and utilize this feature to predict the occupancy results at timestamp $t-k$.
Besides, we design 3D convolutional-based short-term and long-term temporal decoders to predict the pseudo voxel features. 
Furthermore, we propose to use independent temporal decoders and a shared occupancy prediction head in TEOcc.
As a plug-and-play method, the temporal enhancement module can be easily incorporated into existing occupancy prediction methods with additional 0.1$\times$ training time costs.
Extensive experiments on the Occ-3D nuScenes validation dataset show the effectiveness of the proposed TEOcc.
Specifically, TEOcc-RC achieves 42.90 mIoU, outperforming all the previous occupancy networks and achieving new state-of-the-art occupancy prediction results on the leaderboard.







\bibliography{mybibfile}

\end{document}